\documentclass[conference]{IEEEtran}
\IEEEoverridecommandlockouts
\usepackage{url}
\usepackage{cite}
\usepackage{amsmath,amssymb,amsfonts}
\usepackage{algorithmic}
\usepackage{graphicx}
\usepackage{textcomp}
\usepackage{xcolor}
\usepackage{multirow}
\usepackage[hidelinks]{hyperref}
\def\BibTeX{{\rm B\kern-.05em{\sc i\kern-.025em b}\kern-.08em
    T\kern-.1667em\lower.7ex\hbox{E}\kern-.125emX}}
\begin{document}

\title{TransConv-DDPM: Enhanced Diffusion Model for Generating Time-Series Data in Healthcare

\thanks{
This manuscript is an author-prepared version of work published at
\emph{IEEE COMPSAC 2025}. The final version is available at
\url{https://doi.org/10.1109/COMPSAC65507.2025.00114}
}

}

\author{\IEEEauthorblockN{1\textsuperscript{st} Md Shahriar Kabir}
\IEEEauthorblockA{\textit{Department of Computer Science} \\
\textit{Texas State University}\\
San Marcos, TX, USA \\
cpi12@txstate.edu}
\and
\IEEEauthorblockN{2\textsuperscript{nd} Sana Alamgeer}
\IEEEauthorblockA{\textit{Department of Computer Science} \\
\textit{Texas State University}\\
San Marcos, TX, USA \\
sana.alamgeer@txstate.edu}
\and
\IEEEauthorblockN{3\textsuperscript{rd} Minakshi Debnath}
\IEEEauthorblockA{\textit{Department of Computer Science} \\
\textit{Texas State University}\\
San Marcos, TX, USA \\
stg60@txstate.edu}
\and
\IEEEauthorblockN{4\textsuperscript{th} Anne H. H. Ngu}
\IEEEauthorblockA{\textit{Department of Computer Science} \\
\textit{Texas State University}\\
San Marcos, TX, USA \\
angu@txstate.edu}
}

\maketitle

\begin{abstract}

The lack of real-world data in clinical fields poses a major obstacle in training effective AI models for diagnostic and preventive tools in medicine. Generative AI has shown promise in increasing data volume and enhancing model training, particularly in computer vision and natural language processing (NLP) domains. However, generating physiological time-series data, a common type in medical AI applications, presents unique challenges due to its inherent complexity and variability.
This paper introduces TransConv-DDPM, an enhanced generative AI method for biomechanical and physiological time-series data generation. The model employs a denoising diffusion probabilistic model (DDPM) with U-Net, multi-scale convolution modules, and a transformer layer to capture both global and local temporal dependencies. We evaluated TransConv-DDPM on three diverse datasets, generating both long and short-sequence time-series data. Quantitative comparisons against state-of-the-art methods, TimeGAN and Diffusion-TS, using four performance metrics, demonstrated promising results, particularly on the SmartFallMM and EEG datasets, where it effectively captured the more gradual temporal change patterns between data points.
Additionally, a utility test on the SmartFallMM dataset revealed that adding synthetic fall data generated by TransConv-DDPM improved predictive model performance, showing a 13.64\% improvement in F1-score and a 14.93\% increase in overall accuracy compared to the baseline model trained solely on fall data from the SmartFallMM dataset. 
These findings highlight the potential of TransConv-DDPM to generate high-quality synthetic data for real-world applications.

\end{abstract}


\section{Introduction}

There is a growing interest in leveraging physiological and biomechanical measurements, often represented as time-series data from wearable devices, to predict clinically relevant events. Examples include forecasting glucose concentration levels~\cite{he2019causalbg}, assessing susceptibility to cardiac arrhythmias~\cite{Walker2017, Varshneya2021}, and predicting limb movements using EMG (Electromyography) signals~\cite{smirnov2021solving}. Physiological and biomechanical processes are inherently complex and often involve nonlinear dynamics, necessitating the use of advanced deep-learning models for accurate analysis and detection. However, these models typically require large datasets for effective training. With limited data, they are prone to overfitting and struggle to generalize effectively. Moreover, the intricate nature of physiological and biomechanical data often makes it impractical to develop reliable physics-based models, further emphasizing the need for robust data-driven approaches to generate meaningful insights.

An example is the forecasting of falls in stick-balancing on the fingertip~\cite{Marquez2023}. The dynamics of the movements of the balanced stick are exceedingly complex because there is a need to account for human's delayed reaction time and external force. The best mathematical model that approximates the dynamic of balancing sticks took more than a decade to develop~\cite{Cabrera2004b}. Moreover, it is impossible to collect a large amount of real-world stick-balancing data 
to train a model because of the onset of human fatigue in performing a large number of balancing trials. 

Learning the dynamical signatures of stick falls serves as a foundational step for advancing current theories of human balance and fall dynamics~\cite{Insperger2011}. For instance, bi-directional Long Short-Term Memory (LSTM) deep learning models have been used to predict stick falls using data generated by a mathematical stick-balancing model based on an inverted pendulum~\cite{debnath2024pole}. This bi-LSTM model when tested on the mathematically generated data performed very well with accuracy over 95\%, but failed miserably (less than 20\%) when tested with data collected from real human stick-balance. When this model was adapted using transfer learning with a small amount of real human stick-balancing data to predict the falling of a stick, it could predict within the next 2.35 seconds with 60\% - 70\% accuracy.
While this approach demonstrates the potential of combining physics-simulated and real-world data, it underscores a broader challenge: developing accurate physical models or simulators to fully capture the complexities of physiological or biomechanical processes is often highly time-consuming, if not impossible. 

Generative AI has emerged as a promising area of research for addressing such challenges, particularly in creating synthetic data across various domains such as images and text. Techniques such as Generative Adversarial Networks (GANs)~\cite{goodfellow2014}, Variational Autoencoders (VAEs)~\cite{kingma2019}, and Denoising Diffusion Probabilistic Models (DDPMs)~\cite{ho2020denoising} have been successfully employed for synthetic time-series generation. Among these, diffusion models have recently gained popularity, showcasing exceptional performance in diverse applications~\cite{math10152733, REUTOV2023335, mario2024, nwae2762024}. By synthesizing realistic time-series data, these models have the potential to generate abundant and diverse datasets, enabling the training of optimal machine-learning models for a wide range of clinical applications. This observation leads to a pivotal research question: \textit{Can we leverage diffusion-based generative AI tools to generate realistic time-series data (e.g., physiological and biomechanical data) that can address the data scarcity problem in clinical AI applications?}
 
The current architecture of DDPMs has a significant limitation: the standard U-Net architecture struggles to model long-range dependencies, which restricts its ability to generate complex time-series data. 
To address this limitation, we propose an enhanced DDPM, termed TransConv-DDPM, which incorporates two key innovations: (1) A transformer layer positioned between the downsampling and upsampling paths of U-Net to effectively capture long-range temporal dependencies. (2) A multi-scale convolution module designed to extract features at different temporal scales, enhancing the capacity of the model to process data with varying frequencies and lengths. 
The main contributions of our paper are:
\begin{itemize}
    \item An enhanced DDPM model for generating time-series data capable of handling varying sequence lengths and time steps. 
    \item A comparative analysis of TransConv-DDPM with other generative AI-based models for time-series data generation.
    \item A validation of the generated data quality using multiple metrics and diverse datasets.
    \item A demonstration of the utility of synthetic data in training a fall detection model using the combination of real and synthetic fall data.
\end{itemize}
\section{Related Work}
In this section, we review the strengths and limitations of existing methods for time-series data generation, categorizing them into non-generative and generative approaches. 

\subsection{Non-Generative AI-based Techniques}
For time-series data generation, several non-generative approaches have been explored to augment or synthesize data. These methods include mathematical modeling, traditional data augmentation techniques, and pose estimation from video analysis. 

Mathematical Models: Mathematical models have been widely used for generating synthetic time-series data, particularly when data collection is constrained. For instance, the inverted pendulum model~\cite{Cabrera2004b, Insperger2014, Milton2016, Milton2018, Milton2019, Insperger2021, Nagy2023} has been applied to simulate the dynamics of stick-balancing, i.e., an intricate task involving chaotic movements. These models leverage known physical laws to replicate specific behaviors. While they provide a controlled environment for studying well-understood processes, they are often limited by their domain specificity. The development of such models requires deep domain knowledge and extensive effort. 

Data Augmentation Techniques: Traditional augmentation methods, which involve simple transformations of existing data~\cite{augmentTS2017icmi, augts2017icdm}, have been a popular approach due to their ease of implementation and low computational cost. Techniques like jittering, scaling, and time warping introduce variability by adding noise, altering amplitude, or distorting temporal structures. However, these methods often fail to preserve the intricate temporal dependencies and complex patterns inherent in real-world data, particularly chaotic or multivariate time-series. Additionally, unsupervised/arbitrary application of such techniques may introduce unrealistic artifacts, creating spurious correlations that can mislead the downstream models. Advanced augmentation techniques offer more sophisticated solutions, such as Dynamic Time Warping (DTW)~\cite{Sakoe1978} and decomposition-based methods~\cite{Cleveland1990}. These methods are effective in maintaining structural consistency but fail to capture the unpredictability and interdependencies in highly dynamic data~\cite{Wen2021, Iglesias2023}.



Pose Estimation or Video-Based Extraction Techniques: These include alternative approaches that involve extracting time-series data from video recordings through pose estimation techniques~\cite{Singh2022, Dubey2023, Stenum2024}. By analyzing video frames, algorithms can track and quantify human movements over time, generating time-series representations of joint angles, velocities, and other kinematic variables. 
However, their effectiveness is contingent upon the availability of high-quality video data and accurate pose estimation algorithms. Challenges such as occlusions, varying lighting conditions, and individual differences in movement patterns can adversely affect the accuracy and reliability of the extracted time-series data~\cite{Tang2023}. 

\subsection{Generative AI-based Techniques}
Generative AI-based techniques have significantly advanced the generation of time-series data, offering robust tools for modeling complex temporal patterns across various domains.

Generative Adversarial Networks (GANs): GANs have been adapted to time-series data generation by combining adversarial training with supervised learning~\cite{Mogren2016, Xu2020, Xu2020COTGAN, Lin2019GeneratingHS}. These models integrate temporal dynamics through a supervised loss that aligns the latent space with the temporal features of the data, enhancing their ability to produce realistic sequences. Additionally, some of these models, TimeGAN~\cite{Srinivasan2022} using recurrence neural network, and TTS-GAN~\cite{Lin2019} using the transformer architecture have demonstrated the ability to capture long-range dependencies and model complex temporal patterns more effectively. While this integration improves the representational power of GANs, they still face limitations such as mode collapse, where the generator produces a limited diversity of outputs~\cite{bauer2024}, and challenges in fully capturing intricate long-range temporal dependencies due to the use of adversarial training framework.

Variational Autoencoders (VAEs): VAEs offer a probabilistic framework for generating time-series data by learning latent representations~\cite{desai2021, LiHongming2023}. By encoding input sequences into a latent space and decoding them back, VAEs can generate diverse data points by sampling from the latent distribution. However, the assumption of a smooth latent space often limits their ability to generate fine-grained time-series data. They are also less effective in modeling intricate temporal dependencies compared to other models like TimeGAN.

Diffusion Models: Denoising Diffusion Probabilistic Models (DDPMs)~\cite{tian2023ddpm} have gained popularity for generating high-quality time-series data~\cite{10185559, alcaraz2023, suh2024, tian2024}. These models progressively add and remove noise to learn data distributions, resulting in diverse and high-fidelity outputs. While they are particularly strong in capturing complex data distributions, the standard U-Net architecture~\cite{unet2015} often used in diffusion models has limitations in handling long-range dependencies and varying sequence lengths. Recent advancements, such as integrating transformer layers~\cite{sikder2024}, have improved their ability to model long-range dependencies. However, existing architectures often do not fully leverage multi-scale temporal features, which are critical for effectively capturing both local and global temporal dependencies in time-series data. 

Digital Twins: This technology is an emerging method for generating synthetic data, offering precise simulations of physical systems. A recent study~\cite{Tufisi2024} leverages this approach to simulate forward fall dynamics, creating synthetic inertial data for training machine learning models. This method allows the controlled generation of diverse datasets, bypassing the ethical and logistical challenges of real-world data collection. 
However, the approach requires extensive calibration to ensure alignment with real-world variability and struggles to generalize across diverse populations and sensor configurations.

\begin{figure*}[htb!]
  \centering
  \includegraphics[width=0.8\textwidth]{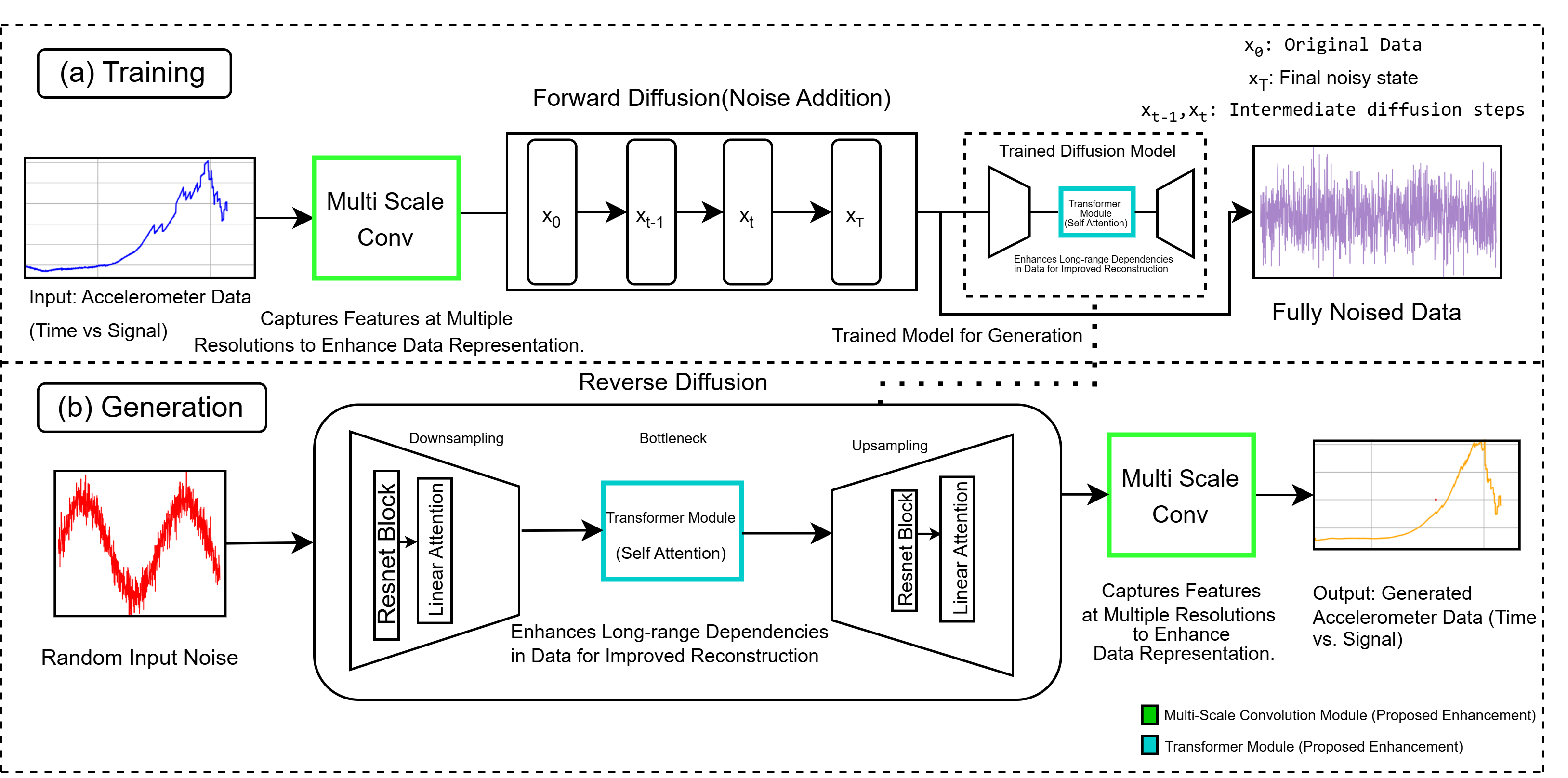}
  \caption{Illustration of the proposed model architecture. (a) Training: The diffusion process adds noise to the input accelerometer data for feature extraction. (b) Generation: The denoising process generates synthetic accelerometer data using a Transformer-enhanced model.}
  \label{fig:my_image}
  \vspace{-10px}
\end{figure*}

\section{Methodology}
This section introduces the proposed TransConv-DDPM framework, a non-autoregressive Denoising Diffusion Probabilistic model (DDPM) designed to generate high-fidelity time-series data. The DDPM process is adapted as-is from \cite{tian2023ddpm} and consists of a forward diffusion process (training) and a reverse denoising process (generation). The primary novelty lies in addressing the limitations of the built-in U-Net architecture by introducing two significant enhancements: (1) a multi-scale convolution module to improve the extraction of localized temporal patterns and (2) a transformer layer to capture global temporal dependencies efficiently. Figure~\ref{fig:my_image} illustrates the overall architecture, highlighting the modified components.

\paragraph{Forward Diffusion (Training)} During training, Gaussian noise is progressively added to the input sequence \( x_0 \) over \( T \) time steps, transforming it into a noisy version \( x_T \). The process is defined as: 
\begin{equation}
    q(x_t | x_0) = \mathcal{N}(x_t; \sqrt{\alpha_t}x_0, (1 - \alpha_t)I),
\end{equation}
where \( \alpha_t \) represents the noise schedule that controls the gradual addition of noise.

\paragraph{Reverse Denoising (Generation)} During generation, the model starts with random Gaussian noise \( x_T \) and iteratively removes noise to reconstruct the original sequence \( x_0 \). The reverse process is parameterized as:
\begin{equation}
    p_\theta(x_{t-1} | x_t) = \mathcal{N}(x_{t-1}; \mu_\theta(x_t, t), \sigma_t^2 I),
\end{equation}
where \( \mu_\theta(x_t, t) \) and \( \sigma_t^2 \) are learned parameters.

The standard U-Net architecture, commonly used in DDPMs, serves as the starting point for the TransConv-DDPM framework. It is designed for hierarchical feature extraction and consists of three primary components:
\begin{itemize}
    \item Downsampling Path: This path extracts features while progressively reducing the temporal resolution using convolutional layers.
    \item Bottleneck: This component processes the lowest-resolution feature map, capturing abstract representations of the input.
    \item Upsampling Path: This path reconstructs the original temporal resolution while integrating features from the downsampling path through skip connections.
\end{itemize}

However, the standard U-Net architecture has limitations when applied to time-series data. Its reliance on localized convolutional operations limits its ability to capture global temporal dependencies, which are critical for modeling sequential patterns across extended time horizons. Additionally, the fixed receptive field of convolutional kernels restricts adaptability to varying temporal resolutions, making it less effective for capturing both fine-grained and broader temporal structures. To address these limitations, the TransConv-DDPM introduces two significant enhancements:
\begin{itemize}
    \item Multi-Scale Convolution Modules: These modules extract features across multiple temporal resolutions, enabling the model to capture localized patterns and adapt to variations in temporal granularity.
    \item Transformer Layer: Positioned at the bottleneck, the transformer layer models global temporal dependencies efficiently through a self-attention mechanism.
\end{itemize}

These modifications enhance the ability of U-Net architecture to balance local and global feature extraction, ensuring the generation of high-quality time-series data.

\subsection{Multi-Scale Convolution for Local Dependencies}
To capture short-term variations such as noise, spikes, or small fluctuations in time-series data, the multi-scale convolution module extracts localized features at multiple temporal resolutions. This is achieved by applying 1D convolutions with multiple kernel sizes and dilation rates:
\begin{equation}
    y_i = \text{Conv1D}(x, k_i, d_i), \quad i \in \{1, 2, \dots, N\},
\end{equation}
where \( x \) is the input sequence, \( k_i \) is the kernel size, \( d_i \) is the dilation rate, and \( y_i \) represents the convolution output at scale \( i \).

The outputs from different scales are dynamically weighted using learnable attention coefficients:
\begin{equation}
    \alpha_i = \frac{\exp(\beta_i)}{\sum_j \exp(\beta_j)}, \quad y_{\text{final}} = \sum_i \alpha_i y_i,
\end{equation}
where \( \beta_i \) are learnable parameters. The final aggregated output is normalized and activated using the SiLU function:
\begin{equation}
    y_{\text{out}} = \text{SiLU}(\text{BatchNorm}(y_{\text{final}})).
\end{equation}

This module effectively captures localized temporal dependencies across varying resolutions.

\subsection{Transformer Layer for Global Dependencies}
While multi-scale convolutions focus on localized patterns, global temporal dependencies, such as trends and long-term relationships, are modeled using a transformer layer. Integrated at the bottleneck of the U-Net, the transformer layer operates on the lowest-resolution feature map, minimizing computational overhead while efficiently capturing global relationships.

The transformer uses a self-attention mechanism to compute pairwise relationships between all time steps:
\begin{equation}
    \text{Attention}(Q, K, V) = \text{softmax}\left(\frac{QK^T}{\sqrt{d_k}}\right)V,
\end{equation}
where \( Q, K, V \) are the query, key, and value matrices, and \( d_k \) is the attention head dimension. To ensure scalability, the number of attention heads is dynamically adjusted:
\begin{equation}
    h_{\text{adaptive}} = \max\left(1, \frac{L}{C}\right),
\end{equation}
where \( L \) is the sequence length and \( C \) is a scaling factor.

The transformer stabilizes training through a residual connection:
\begin{equation}
    z_{\text{output}} = \lambda \cdot \text{Attention}(x) + x,
\end{equation}
where \( \lambda \) is a learnable scaling factor. By integrating the transformer with multi-scale convolution modules, the framework effectively balances local and global feature extraction.


\section{Experimental Setup} 

\subsection{Datasets}
We evaluated the effectiveness of the TransConv-DDPM model using three datasets: Human Stick-balancing~\cite{debnath2024pole}, 
SmartFallMM~\footnote{https://github.com/txst-cs-smartfall/SmartFallMM-Dataset}, and EEG~\cite{shoeb2009}.
Those datasets constitute both biomechanical and physiological data which are difficult to collect in large quantities in the real world.
Human Stick-balancing dataset was obtained from an external source.
Stick-balancing at the fingertip was performed by subjects sitting in a chair while facing a blank black screen. The stick 
has reflective markers attached at each end. 
A high-speed motion capture system 
was used to record the position of reflective markers
during stick balancing as a function of time. Data is recorded  at 250 Hz. Balance trials in which the fall occurs in
the sagittal plane were processed and stored as CSV files.
The dataset contains six features: time ($t$), angle of the stick ($\phi$), angular velocity of the stick ($\phi'$), horizontal position of the stick base ($X_1$), velocity of the stick base ($X_1'$), and external forces applied to the stick base ($F$). Each file represents a trial. Different trials can contain different variable numbers of data points. A total of 40 files were used for this study. 

SmartFallMM is a multi-modal human activity dataset collected under IRB 9461. 
The dataset includes activity data from 23 older and 19 younger adults. Each participant wore four IMU sensors (smartwatch, smartphone two meta sensors) positioned at the wrists and hips allowing for the collection of data from four crucial joints of the human body during various movements. Younger participants were instructed to simulate five types of falls (front, back, left, right, and rotational) on an air mattress with five repetitions each. 
We used 3D accelerometer data (X, Y, Z axes) of the five types of fall data collected from the smartwatch placed on the left wrist of the participants for data generation. In total, we utilized 237 files (from younger participants) containing only fall data, with each file comprising 240 data points sampled at 32 ms intervals. 

The EEG data we used is a publicly available dataset called Chb-Mit \cite{Guttag2010}. This dataset, collected at Children’s Hospital in Boston, comprises EEG recordings from pediatric subjects (aged 1.5–22) with intractable seizures monitored for several days following withdrawal of anti-seizure medication to evaluate surgical candidacy. In this study, we selected seizure data from 10 participants (ranging from ch01 to ch10). 

\subsection{Data Preparation}
We downsampled the EEG data from 256 Hz to 100 Hz to reduce computational complexity. Following downsampling, we selected the middle 900 data points from each seizure file. This approach is commonly adopted in time-series analysis to mitigate edge effects and ensure that the selected segment is representative of the overall signal, as the central portion often contains more stable and relevant information. Then, from the downsampled data, we used five pairs of EEG channels, ranked using SHAP analysis~\cite{shap2017}: FP1-F7, F7-T7, T7-P7, P7-O1, and FP1-F3. Next, we split the selected data points into non-overlapping windows of size 128, with each sequence having the same length of 128 to maintain uniformity for training. For the SmartFall dataset, we created non-overlapping windows of size 240 by taking all data points from each file containing fall data. We ensured that each window retained the sequence length of 240.
For the Human Stick-balancing dataset, the data samples come with variations of different lengths.  We chose a fixed length of 2,576 
based on the average duration across all samples.
A significant change in signal pattern always occurs towards the end of the trial when the stick falls.  We thus only select 2,576 data points from the end of each sample to use as input.
For each sample, all six features were included in the training process. 
Similar to the SmartFallMM dataset, we used the entire set of samples to create non-overlapping windows with a size equal to the length of the data points.

\subsection{Training and Testing Parameters}
We trained our TransConv-DDPM model using the PyTorch framework.
Across all datasets, we used a consistent learning rate of 8e-5 and a batch size of 32.
However, the number of training iterations varied depending on the specific characteristics of each dataset. For the EEG dataset, we trained the model for 8,000 iterations because the data contains complex seizure-related signals with multiscale temporal patterns and noise, requiring longer training to capture effectively. The SmartFallMM dataset converged well after 5,000 iterations due to its shorter sequences (240 time-steps) and smoother patterns related to fall dynamics. For the Human Stick-balance dataset, which has much longer sequences (2,576 time-steps) and more chaotic dynamics involving angular velocity and external forces, the model required 7,000 iterations at least to learn the intricate temporal dependencies. 

We also adjusted the channel dimensions for each dataset to match their specific characteristics. The SmartFallMM dataset uses three features (X, Y, Z axes), but the variability in movement patterns justified using a larger channel dimension of 128 to better capture these nuances. For the EEG dataset, we selected a channel dimension of 64, which was sufficient for capturing the relationships between the SHAP-ranked electrode pairs while avoiding overfitting due to the relatively small feature space. Similarly, the Human Stick-balance dataset, which includes six features (like angle, velocity, and force), using a channel dimension of 64 balanced the need for computational efficiency with model capacity. In all cases, the model utilized dimension multipliers of (1, 2, 4, 8), which allowed for a gradual increase in feature extraction complexity across layers.

Since the proposed method relies on unconditional data generation and represents an example of unsupervised learning, it does not involve a conventional train-test split during the training process.  
However, to ensure the quality and reliability of the generated data, we reserved a portion of the original dataset and used it to estimate the quality of the generated samples using the aforementioned quality evaluation metrics. 
After training with each dataset, we generated 1,000 synthetic sequences for each dataset with varying lengths (2576, 240, 128 for Human Stick-balancing, SmartfallMM, and EEG datasets, respectively), reconstructed them to their original scale, and saved the outputs for evaluation. 

\section{Results}

This section presents the qualitative and quantitative evaluation of the data generated by TransConv-DDPM.
Evaluating the quality of time-series data is inherently challenging, as there is no single definitive way to determine whether the generated data perfectly matches the real data. To address this, we employ four metrics to assess the alignment between real and synthetic data in terms of both overall trends and finer temporal details. These evaluations highlight the ability of the model to generate realistic and coherent time-series data that closely mirrors the characteristics of the real datasets.

\begin{figure*}[ht!]
    \centering
    \includegraphics[width=0.95\textwidth]{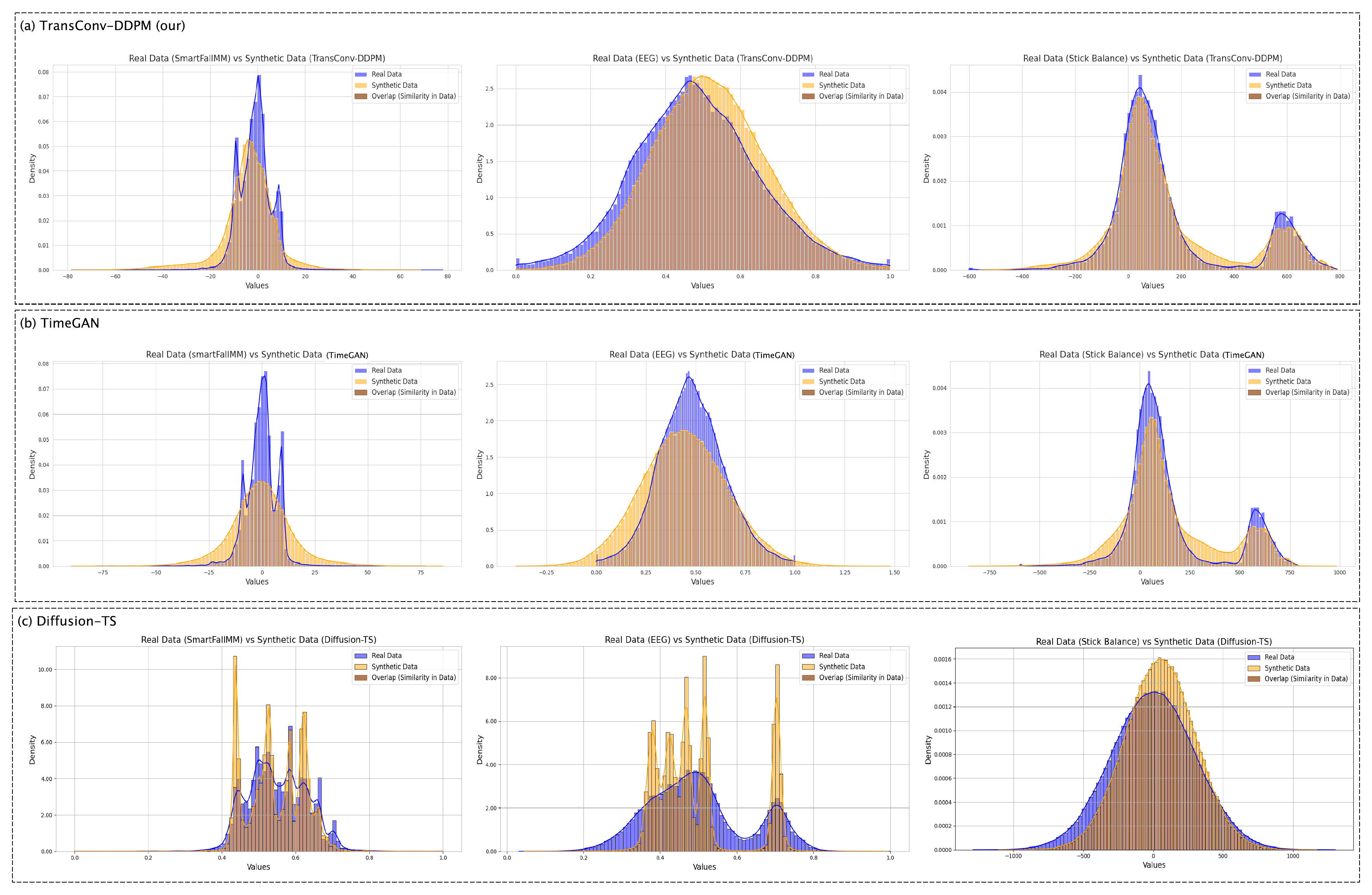}
    \caption{Distribution comparison for the SmartFallMM, EEG, and Stick-balance datasets with synthetic data generated from (a) TransConv-DDPM (our), (b) TimeGAN, and (c) Diffusion-TS methods.}
    \label{fig:viz_real_vs_fake_data}
    \vspace{-10px}
\end{figure*}

\subsection{Qualitative Analysis: Distribution Comparison}
We first assess the quality of the synthetic data generated by TransConv-DDPM through data distribution comparisons using kernel density estimation (KDE) plots. Figures~\ref{fig:viz_real_vs_fake_data}(a), (b), and (c) illustrate the distribution of real and generated data for the SmartFallMM, EEG, and Human Stick-balance datasets (from left to right in each row), across three methods TransConv-DDPM, TimeGAN, and Diffusion-TS, respectively. 

\textbf{SmartFallMM: Real vs Generated Data:} In Figure~\ref{fig:viz_real_vs_fake_data}(a), the left KDE plot for TransConv-DDPM shows a high degree of overlap between the real data (blue) and synthetic data (orange), with significant alignment in high-density regions as represented by the brown areas. The model shows a strong performance in regions where the real data exhibits sharp peaks, reflecting its ability to generate realistic data. This indicates that the proposed model effectively captures the underlying statistical distribution of the SmartFallMM dataset. In Figure~\ref{fig:viz_real_vs_fake_data}(b), the left KDE plot for TimeGAN demonstrates less alignment between the real and synthetic data compared to TransConv-DDPM. The synthetic data distribution appears broader, with weaker overlap in high-density regions and significant divergence in the tails of the distribution. 
This suggests that TimeGAN struggles to capture the core statistical properties of the SmartFallMM dataset, resulting in less realistic synthetic data generation.
For Diffusion-TS, the left KDE plot in Figure~\ref{fig:viz_real_vs_fake_data}(c), shows a moderate degree of overlap between the real and synthetic data. While the model captures certain high-density regions reasonably well, the alignment is fragmented, with several sharp peaks in the synthetic data that do not correspond to those in the real data. This suggests that Diffusion-TS performs better than TimeGAN but falls short of the level of alignment achieved by TransConv-DDPM. Overall, TransConv-DDPM demonstrates the best 
performance for SmartFallMM.

\textbf{EEG: Real vs Generated Data:} For synthetic EEG data generated from TransConv-DDPM, the middle KDE plot in Figure~\ref{fig:viz_real_vs_fake_data}(a), demonstrates the best alignment between real data (blue) and synthetic data (orange), with significant overlap in high-density regions. The proposed model effectively captures the statistical properties of the real data, particularly in the central range of the distribution, reflecting its ability to generate realistic synthetic EEG data. In comparison, TimeGAN (middle KDE plot in Figure~\ref{fig:viz_real_vs_fake_data}(b) shows moderate alignment, with the synthetic data distribution being narrower and less aligned with the real data in the tails of the distribution. While the model captures the general structure of the real data, the reduced overlap in high-density regions indicates weaker performance in replicating the full statistical distribution of the EEG data. Diffusion-TS (bottom KDE plot in Figure~\ref{fig:viz_real_vs_fake_data}(c)) displays the weakest performance among the three models. The synthetic data distribution is highly fragmented, with sharp peaks that do not align well with the real data. The lack of smooth overlap in key regions highlights the challenges faced by Diffusion-TS in capturing the continuous and rhythmic patterns characteristic of EEG data. Overall, TransConv-DDPM achieves the highest fidelity to the real EEG data, outperforming both TimeGAN and Diffusion-TS.
\\

\textbf{Human Stick-balance: Real vs Generated Data:}
For this dataset, the right KDE plot in Figure~\ref{fig:viz_real_vs_fake_data}(a), shows that the TransConv-DDPM model achieves a high degree of alignment between the real data (blue) and synthetic data (orange) in the main high-density region around the central peak. This indicates that the proposed model captures the core statistical properties of the data effectively. However, slight discrepancies are observed in the secondary peak and tails of the distribution, suggesting challenges in modeling the variability and outlier patterns present in this dataset. 
In comparison, TimeGAN (middle KDE plot in Figure~\ref{fig:viz_real_vs_fake_data}(b)) demonstrates a broader synthetic distribution with weaker alignment in the secondary peak and noticeable deviations in the central region. While there is some overlap, the synthetic data fails to replicate the sharpness and specificity of the real data, indicating that TimeGAN struggles to capture the more chaotic temporal patterns of the Human Stick-balance dataset. 
The right KDE plot in Figure~\ref{fig:viz_real_vs_fake_data}(c) shows that the synthetic data generated by Diffusion-TS aligns well with the real data in the central high-density region. The overlap (brown areas) between the real data (blue) and synthetic data (orange) indicates that the model captures the core statistical properties of the dataset. Overall, Diffusion-TS achieves the highest alignment between real and synthetic data for the Human Stick-balance dataset. 

\subsection{Quantitative Analysis: Performance Metrics}
\label{sec:perMetrics}
\begin{table*}[ht!]
\centering
\caption{Quantitative results across three datasets. Bold values represent the best performance (lowest score $\downarrow$) for each metric and dataset, highlighting the comparison with the proposed TransConv-DDPM method.}
\label{tab:quantitative_results}
\resizebox{0.67\textwidth}{!}{%
\begin{tabular}{|c|c|c|c|c|}
\hline
\multirow{2}{*}{\textbf{Evaluation Metrics}} & \multirow{2}{*}{\textbf{Methods}} & \multicolumn{3}{c|}{\textbf{Datasets}} \\ \cline{3-5}
 &     & \textbf{Stick-balance} & \textbf{SmartFallMM} & \textbf{EEG} \\ \hline
\multirow{3}{*}{Context-FID Score ($\downarrow$)} & TransConv-DDPM (ours)  & 2.7551  & \textbf{0.4097}  & \textbf{1.7532} \\  
                            & TimeGAN          & 2.1320  & 0.5318  & 2.4253         \\ 
                            & Diffusion-TS     &   \textbf{1.5312}  & 0.7220 & 5.1992    \\ \hline
\multirow{3}{*}{Discriminative Score ($\downarrow$)} & TransConv-DDPM (ours)  & 0.5215 & 0.7042 & \textbf{0.3004} \\ 
                            & TimeGAN          & 0.6608        &  0.8543         & 0.4858        \\ 
                            & Diffusion-TS  & \textbf{0.4834}   &  \textbf{ 0.5000}  &  0.3214  \\ \hline
\multirow{3}{*}{Predictive Score ($\downarrow$)} & TransConv-DDPM (ours) & \textbf{0.5246} & 0.7352  & 0.7453 \\  
                            & TimeGAN          & 0.6839  & \textbf{0.6642}       & \textbf{0.7060}        \\ 
                            & Diffusion-TS     & 0.5843 & 0.7736 &  0.9762      \\ \hline
\multirow{3}{*}{JSD ($\downarrow$)}       & TransConv-DDPM (ours) & \textbf{0.0009} & \textbf{0.0567}  & \textbf{0.0108} \\  
                            & TimeGAN          & 0.0802        & 0.1254         & 0.0120       \\ 
                            & Diffusion-TS     &   0.0412 & 0.2413 & 0.4483  \\ \hline
\end{tabular}%
}
\vspace{-10px}
\end{table*}

To further evaluate the quality of the synthetic time-series data generated by TransConv-DDPM, we compared its performance with two state-of-the-art models, TimeGAN~\cite{timegan2019} and Diffusion-TS~\cite{diffusionts2024}, across three datasets using multiple performance metrics. Since our focus is to analyze how effectively the generated data aligns with the statistical properties of real data, we, therefore, choose the following metrics for performance evaluation:
\begin{itemize}
    \item \textbf{Context-Fidelity (FID) Score:} Measures the quality of the synthetic time-series samples by calculating the difference in time-series representations within a local contextual embedding space. Lower FID scores signify better quality~\cite{paul2022psa}.
    \item \textbf{Discriminative Score:} Evaluates the similarity between real and synthetic data using a classification model trained to distinguish the two data distributions, with lower scores indicating greater similarity~\cite{yoon2019time}.
    \item \textbf{Predictive Score:} Assesses the usefulness of the synthetic data for predictive tasks by training a sequence model using the train-synthesis-and-test-real (TSTR) approach to predict next-step temporal vectors. Lower predictive scores indicate higher utility of synthetic data for predictive tasks~\cite{yoon2019time}.
    \item \textbf{Jensen-Shannon Divergence (JSD):} Measures the probability distributions of real and generated data alignment, with smaller values indicating greater similarity~\cite{jsd1997}.
\end{itemize}

We summarized the quantitative results for these metrics in Table~\ref{tab:quantitative_results}. For the Human Stick-balance dataset, Diffusion-TS achieves the best performance in Context-FID Score (1.5312) and Discriminative Score (0.4834), reflecting its strength in aligning with the statistical properties of the real data and generating synthetic data that is harder to differentiate. However, TransConv-DDPM outperforms in Predictive Score (0.5246) and JSD (0.0009), highlighting its ability to better preserve the overall data distribution and support predictive tasks. These observations align with the KDE plots, where TransConv-DDPM exhibits strong overlap in the central high-density region and captures secondary peaks more effectively than TimeGAN. 

For the SmartFallMM dataset, TransConv-DDPM achieves the best performance in Context-FID Score (0.4097) and JSD (0.0567), demonstrating its ability to preserve the statistical context and align closely with the overall data distribution. Diffusion-TS, however, achieves the best Discriminative Score (0.5000), while TimeGAN achieves the best Predictive Score (0.6642). These results highlight the strengths of Diffusion-TS and TimeGAN in task-specific metrics. These results align with the observations from the KDE plots, where TransConv-DDPM showed the highest overlap with the real data distribution, while Diffusion-TS and TimeGAN exhibited more fragmented or broader distributions, limiting their performance on overall alignment metrics. Despite this, TransConv-DDPM maintains its competitive edge by excelling in metrics that reflect fidelity to the statistical properties of the dataset.

For the EEG dataset, TransConv-DDPM demonstrates the best performance in Context-FID Score (1.7532), Discriminative Score (0.3004), and JSD (0.0108), showcasing its superior ability to preserve context, align closely with the overall data distribution, and generate data that is harder to differentiate from real data. These results align with the KDE plots, where TransConv-DDPM exhibits significant overlap between the real and synthetic data, particularly in high-density regions. However, TimeGAN achieves the best Predictive Score (0.7060), highlighting its advantage in supporting predictive tasks. This is consistent with its KDE plot, which shows reasonable alignment in key patterns but lacks the precise overlap observed with TransConv-DDPM, limiting its overall fidelity to the real data.

In summary, TransConv-DDPM demonstrates strong performance across the three datasets, excelling particularly in the SmartFallMM and EEG datasets due to its ability to capture gradual changes in temporal dependencies. 
In the EEG dataset, despite all methods utilizing the same SHAP-ranked channels, TransConv-DDPM effectively modeled seizure-related brain activity, which contains meaningful temporal features critical for distinguishing between classes. In contrast, the Human Stick-balance dataset posed the most challenge to TransCon-DDPM due to the unpredictable dynamics of the balancing process and high variability in external forces and angular velocities. These intricacies made it more challenging to capture fine-grained temporal patterns, resulting in slightly lower performance on some metrics. However, TransConv-DDPM demonstrates strong capabilities in handling gradual temporal patterns and highlights the potential for improvement in modeling dynamic behaviors.


\subsection{Utility Results}
\begin{table*}[!htb]
    \centering
    \caption{Performance comparison of LSTM model trained with real and synthetic data on the SmartFallMM (SMM) dataset. Bold values show the highest gain in the fall detection task.}
    \label{tab:synthetic_data_results}
    \resizebox{0.8\textwidth}{!}{%
    \begin{tabular}{|l|c|c|c|c|c|}
        \hline
        \textbf{Training Data} & \textbf{Precision} & \textbf{Recall} & \textbf{F1-Score} & \textbf{Accuracy} & \textbf{ROC-AUC} \\ \hline
        SMM + Real Fall Data (baseline) & 0.69 & 0.65 & 0.66 & 0.67 & 0.77 \\ \hline
        SMM + TransConv-DDPM (our)  & \textbf{0.76 (+10.14\%)} & \textbf{0.77 (+18.46\%)} & \textbf{0.75 (+13.64\%)} & \textbf{0.77 (+14.93\%)} & \textbf{0.85 (+10.39\%)} \\ \hline
        SMM + TimeGAN & 0.71 (+3.77\%) & 0.72 (+11.08\%) & 0.70 (+6.67\%) & 0.69 (+3.28\%) & 0.82 (+6.49\%) \\ \hline
        SMM + Diffusion-TS & 0.744 (+7.83\%) & 0.674 (+3.69\%) & 0.68 (+3.03\%) & 0.694 (+3.58\%) & 0.822 (+6.75\%)   \\ \hline
    \end{tabular}
    }
    \vspace{-10px}
\end{table*}
To evaluate the utility of synthetic data, we conducted four experiments to train and test a fall detection model, assessing whether synthetic fall data can improve the detection performance when added to the original dataset with both ADL and fall samples. For this purpose, we used a Long Short-Term Memory (LSTM) classification model with the SmartFallMM (SMM) dataset, because it showed the highest Context-FID score by the TransConv-DDPM method. We evaluated the performance of the model using accuracy, Area Under the Curve Receiver Operating Characteristic - Area Under the Curve (ROC-AUC), and the macro averages of precision, recall, and F1 score.

In the first experiment, we established a baseline by training the LSTM model on real data comprising 60\% ADL samples and 40\% fall samples.
The data was prepared using overlapping windows of size 128 × 3 with a step size of 10, and split into 60\% training, 20\% validation, and 20\% testing, ensuring subject independence by assigning data from two subjects each to validation and testing. 
In the second, third, and fourth experiments, we expanded the training set by incorporating 20\% synthetic fall samples generated by TransConv-DDPM, TimeGAN, and Diffusion-TS, respectively. 
The expanded training set contained 60\% ADL, 20\% real fall, and 20\% synthetic fall samples. 
To ensure robustness, we repeated training and testing over five iterations, shuffling both the subjects and the synthetic data samples in each iteration, i.e., randomly selecting 20\% synthetic samples from the 1,000 generated samples to avoid overfitting. In Table~\ref{tab:synthetic_data_results}, we have reported the average results.

We designed an LSTM-based classification model with a three-layer architecture: (a) an LSTM layer with 128 units, (b) a dense layer with 128 neurons, ReLU activation, and batch normalization, and (c) an output layer with a sigmoid activation for binary classification. The model was compiled with binary cross-entropy loss and Adam optimizer. The training was done with a batch size of 64 for up to 250 epochs, with early stopping (patience of 20 epochs) to prevent overfitting. We applied a threshold of 0.5 to the output class probabilities, classifying outputs greater than or equal to this value as falls. 

As observed in Table~\ref{tab:synthetic_data_results}, the LSTM model trained with a combination of real and synthetic fall data generated by the TransConv-DDPM method demonstrated the highest performance across all metrics, with +10.14\% improvement in precision, +18.46\% in recall, +13.64\% in F1-score, +14.93\% in accuracy, and +10.39\% in ROC-AUC. The overall increase in F1-score shows that the synthetic data generated by TransConv-DDPM preserves critical statistical properties of real fall data, which enabled the model to converge effectively and accurately distinguish between fall and no-fall samples.
In comparison, the LSTM model trained with synthetic fall data from TimeGAN achieved moderate improvements over the baseline, with +3.77\% improvement in precision, +11.08\% in recall, +6.67\% in F1-score, +3.28\% in accuracy, and +6.49\% in ROC-AUC. While TimeGAN demonstrated some ability to improve model performance, its synthetic data did not achieve the same level of enhancement as TransConv-DDPM. The model trained with synthetic fall data from Diffusion-TS achieved smaller gains across all metrics, with +7.83\% improvement in precision, +3.69\% in recall, +3.03\% in F1-score, +3.58\% in accuracy, and +6.75\% in ROC-AUC. While the gains indicate that Diffusion-TS can contribute to predictive performance, it fell short compared to TransConv-DDPM and performed comparably to TimeGAN in some metrics (recall and F1-score). These results demonstrate that TimeGAN and Diffusion-TS struggled to differentiate between ADL and fall samples when trained with synthetic fall data, highlighting their limitations.

\begin{table}[!ht]
\centering
\caption{Ablation Study Results on Stick-balance Dataset}
\label{tab:ablation_results}
\resizebox{\columnwidth}{!}{%
    \begin{tabular}{|c|c|c|}
        \hline
        \textbf{Metric}          & \textbf{Configuration}    & \textbf{Stick-balance} \\ \hline
        \multirow{4}{*}{Context-FID ($\downarrow$)} 
        & Baseline DDPM             & 3.2554                  \\  
        & + Transformer             & 3.1408                  \\  
        & + Multi-Scale Convolution & 3.0042                  \\  
        & Full Model (Ours)         & \textbf{2.7551}                  \\ \hline
        \multirow{4}{*}{Discriminative Score ($\downarrow$)} 
        & Baseline DDPM             & 0.7740                  \\  
        & + Transformer             & 0.6618                  \\  
        & + Multi-Scale Convolution & 0.7207                  \\  
        & Full Model (Ours)         & \textbf{0.5215}                  \\ \hline
        \multirow{4}{*}{Predictive Score ($\downarrow$)}       
        & Baseline DDPM             & 0.7412                  \\  
        & + Transformer             & 0.6610                  \\  
        & + Multi-Scale Convolution & 0.6534                  \\  
        & Full Model (Ours)         & \textbf{0.5246}                  \\ \hline
        \multirow{4}{*}{JSD ($\downarrow$)}       
        & Baseline DDPM             & 0.0452                  \\  
        & + Transformer             & 0.0084                  \\  
        & + Multi-Scale Convolution & 0.0045                  \\  
        & Full Model (Ours)         & \textbf{0.0009}                  \\ \hline
    \end{tabular}
}
\end{table}
\subsection{Ablation Study}
To assess the impact of architectural advancements in TransConv-DDPM, we conducted an ablation study on four model configurations. These configurations include:
\begin{itemize} 
    \item \textbf{Baseline DDPM:}  the standard model without any modifications. 
    \item \textbf{DDPM + Transformer:} includes a transformer layer to capture long-term temporal dependencies. 
    \item \textbf{DDPM + Multi-Scale Convolution:} adds multi-scale convolution modules to improve feature extraction across multiple temporal scales. 
    \item \textbf{Full Model (TransConv-DDPM):} combines both the transformer layer and multi-scale convolution modules to accurately handle local and global temporal relationships.
\end{itemize}

The results, summarized in Table~\ref{tab:ablation_results}, demonstrate the performance of each configuration on the Human Stick-balance dataset across four evaluation metrics: Context-FID, Discriminative Score, Predictive Score, and JSD. The key observations are as follows:
\begin{itemize} 
    \item \textbf{Baseline DDPM:} The baseline model achieves moderate performance across all metrics, demonstrating its capability to handle basic time-series generation but with notable limitations in modeling complex dependencies (e.g., Context-FID = 3.2554, Discriminative Score = 0.7740). 
    \item \textbf{Impact of Transformer:} Incorporating the transformer layer significantly improves the ability to capture long-range dependencies, leading to a reduction in Context-FID (3.1408) and Discriminative Score (0.6618). 
    \item \textbf{Impact of Multi-Scale Convolution:} Adding multi-scale convolution modules further enhances the performance of the model, particularly in extracting fine-grained temporal features. The Context-FID improves to 3.0042, and the JSD decreases to 0.0045. 
    \item \textbf{Full Model (TransConv-DDPM):} Combining the transformer and multi-scale convolution modules provides optimal performance across all measures. The Context-FID is lowered to 2.7551, Discriminative Score to 0.5215, and JSD to 0.0009, demonstrating the synergy of these two innovations in capturing local and global temporal dependencies. 
\end{itemize}

These findings demonstrate the importance of combining transformer layers and multi-scale convolution modules to obtain cutting-edge performance in time-series data generation. The TransConv-DDPM model displays a greater capacity to balance local and global dependencies, resulting in considerable gains across all evaluation metrics.

\section{Conclusion}
This paper presents TransConv-DDPM, an enhanced denoising diffusion probabilistic model incorporating transformers and multi-scale convolution modules to generate realistic time-series data. Through extensive experiments on three diverse datasets, Human Stick-balance, SmartFallMM, and EEG, the model demonstrates its superiority over state-of-the-art methods such as TimeGAN and Diffusion-TS across four quantitative metrics.

The results validate the ability of the proposed model to capture both local and global temporal dependencies, producing high-quality synthetic data that closely aligns with real-world data. Additionally, we tested the utility of the generated synthetic dataset by training an LSTM-based classifier to predict falls using the SmartFallMM dataset. The classifier trained with a combination of real and synthetic fall data from TransConv-DDPM achieved superior performance, showing a 13.64\% improvement in F1-score and a 14.93\% increase in overall accuracy compared to the baseline model trained solely on real fall data. 
To ensure the robustness of this evaluation and mitigate the risk of overfitting on limited data, we employed 5-fold cross-validation during classifier training. All reported results are averaged across folds to provide a more reliable estimate of generalization performance. This validation strategy reinforces the credibility of the observed improvements and adheres to best practices for evaluating predictive models in healthcare applications.
 also significantly outperformed models trained with synthetic data generated by TimeGAN and Diffusion-TS. 
The ablation study further confirmed the critical role of the transformer and multi-scale convolution modules in improving the performance of the TransConv-DDPM.

By addressing the challenges of time-series data generation, TransConv-DDPM contributes to advancing the field of generative AI. Its potential applications span domains such as healthcare, finance, and industrial monitoring, where realistic synthetic data is crucial for training robust predictive models. 

As part of future work, we will address the challenges posed by the chaotic nature of the Human Stick-balance dataset, where sudden and significant changes in data values make it difficult to predict when the stick will fall. We aim to enhance the model’s ability to capture such dynamic and unpredictable patterns. Additionally, we plan to extend the model’s capabilities to support variable-length sequences, multi-modal data generation, and conditional generation based on contextual information. To further assess generalizability and reliability, we will incorporate larger public healthcare benchmarks (e.g., MIMIC-III, PhysioNet, and KFall), enabling more standardized and rigorous comparisons in clinical AI applications.

\bibliographystyle{unsrt}
\bibliography{ref}

\end{document}